\documentclass[twoside,11pt]{article}

\usepackage[abbrvbib, preprint]{jmlr2e}

\usepackage{amsmath}
\usepackage{mathtools}
\usepackage{booktabs}

\usepackage{algorithm}
\usepackage{algorithmicx}
\usepackage{algpseudocode}

\floatname{algorithm}{Algorithm}

\newtheorem{problem}{Problem}

\ShortHeadings{Unraveling the Black Box of Neural Networks}{Shengjian Chen}

\begin{document}

\title{
  Unraveling the Black Box of Neural Networks:\\
  A Dynamic Extremum Mapper}

\author{\name Shengjian Chen \email {chshengj@alumni.sysu.edu.cn}\\
       \addr Intelligent Robotics Center \email {chensj@jihualab.ac.cn}\\
       Jihua Laboratory\\
       Foshan, 528200, China}

\maketitle

\begin{abstract}
We point out that neural networks are not black boxes, and their generalization stems from the ability to dynamically map a dataset to the extrema of the model function. We further prove that the number of extrema in a neural network is positively correlated with the number of its parameters. We then propose a new algorithm that is significantly different from back-propagation algorithm, which mainly obtains the values of parameters by solving a system of linear equations. Some difficult situations, such as gradient vanishing and overfitting, can be simply explained and dealt with in this framework.
\end{abstract}


\section{Introduction}
Although artificial intelligence models based on neural networks have been extensively studied and widely applied, and their prediction accuracy in fields such as image recognition, natural language processing, text processing and question answering far exceeds that of traditional machine learning algorithms, there is a lack of relevant research on their underlying principles, and they are still generally regarded as black boxes. With the rapid increase of model parameters, from ANN to CNN, RNN, and then to GPT and LLM \citep{wu2025survey}, its complexity also increases sharply, while the stability of the system becomes more vulnerable accordingly. If the model malfunctions, it is impossible for us to quickly identify the root cause of the problem and solve it immediately without understanding its logic. For some fields with low requirements for real-time performance, such as image classification and AI-generated artwork, the application of neural network algorithms can be confidently promoted. However, for some fields that require high real-time performance, especially safety, such as autonomous driving \citep{kiran2021deep}, it is necessary to pay more attention to the underlying principles of neural networks and clarify the conditions under which they take effect and fail, so that artificial intelligence can better serve human society.

Although the model structure of neural networks has become prohibitively complex, some scholars still strive to explore their working principles. \cite{buhrmester2021analysis} investigated the explainers that have been popular in recent years. This method attempts to explain neural networks by analyzing the connections between inputs and outputs. The characteristic of black-box explainers is that it does not need to access the internal structure of the model to reveal all the interaction details of the model. They are mainly divided into ante-hoc systems with a global, model-agnostic feature \citep{lipton2018mythos} and post-hoc ones with a local, model-specific feature \citep{ribeiro2016should}. \cite{oh2019towards} analyzed neural networks from the perspective of reverse engineering, and found that they are extremely vulnerable to different types of attacks, and pointed out that the boundary between a white box and a black box is not obvious. \cite{tishby2015deep} took a different approach and proposed the Information Plane. Furthermore, they believed that the main goal of neural networks was to optimize the Information Bottleneck between the compression and prediction of each layer. \cite{shwartz2017opening} further proved the effectiveness of this method. These works provide useful references for the underlying research of neural networks, but there is still a long way to go.

Current researchers seem to be overly obsessed with using engineering methods to explain how neural networks work, while neglecting to explain from a theoretical or mathematical perspective. Neural networks may seem complex, but their structure is clear and they are basically composed of neurons with the same construction, making them particularly suitable for mathematical analysis. It is necessary to revisit the pioneering work  of \cite{cybenko1989approximation} and \cite{hornik1989multilayer} in this area, which proved that feedforward neural networks can approximate any continuous function on a compact set. That is, a feedforward neural network that adopts a single hidden layer with a sufficient number of neurons, and uses the sigmoid function as the activation function can approximate any complex function with arbitrary accuracy, providing a basic mathematical principle for neural networks. The only drawback of this work is that it does not provide a method on how to find the specific function on a given dataset and whether this function is the optimal one. Our work makes up for its deficiencies.

Specifically, our contributions mainly include the following aspects:

1) We present the main characteristics of an ideal machine learning model, and based on which we provide the general model training steps. This work is mainly discussed in Section \ref{sec: Ideal Model}.

2) We discuss whether neural networks satisfy the ideal model characteristics and point out from a mathematical perspective that neural networks mainly achieve generalization by mapping a dataset to the local extrema of the function. We further present a model training algorithm different from the back-propagation (BP) algorithm, namely the extremum-increment (EI) algorithm. This work is mainly discussed in Sections \ref{sec: neural network} and \ref{sec: EI algorithm}.

3) Based on EI algorithm, we can relatively easily point out the causes of some common problems, such as vanishing/exploding gradients, overfitting, etc., and provide corresponding solutions. This work is mainly discussed in Section \ref{sec: deductions}.

\section{General Characteristics of an Ideal Model}\label{sec: Ideal Model}
Let's temporarily put aside the concept of neural networks and imagine the basic characteristics of a model that satisfies a dataset and the target task. The training goal of machine learning is to obtain a function curve that can precisely fit the inputs of all samples with their corresponding outputs. That is to say, this model can clearly tell us what the exact value of each input sample that is after processing. For example, for classification problems, this model can give an output of ``This is a cat.'' instead of a vague answer of ``This is very likely a cat.''.

\subsection{Precise Mapping}
\begin{figure*}[h]
  	\includegraphics[scale=1.7]{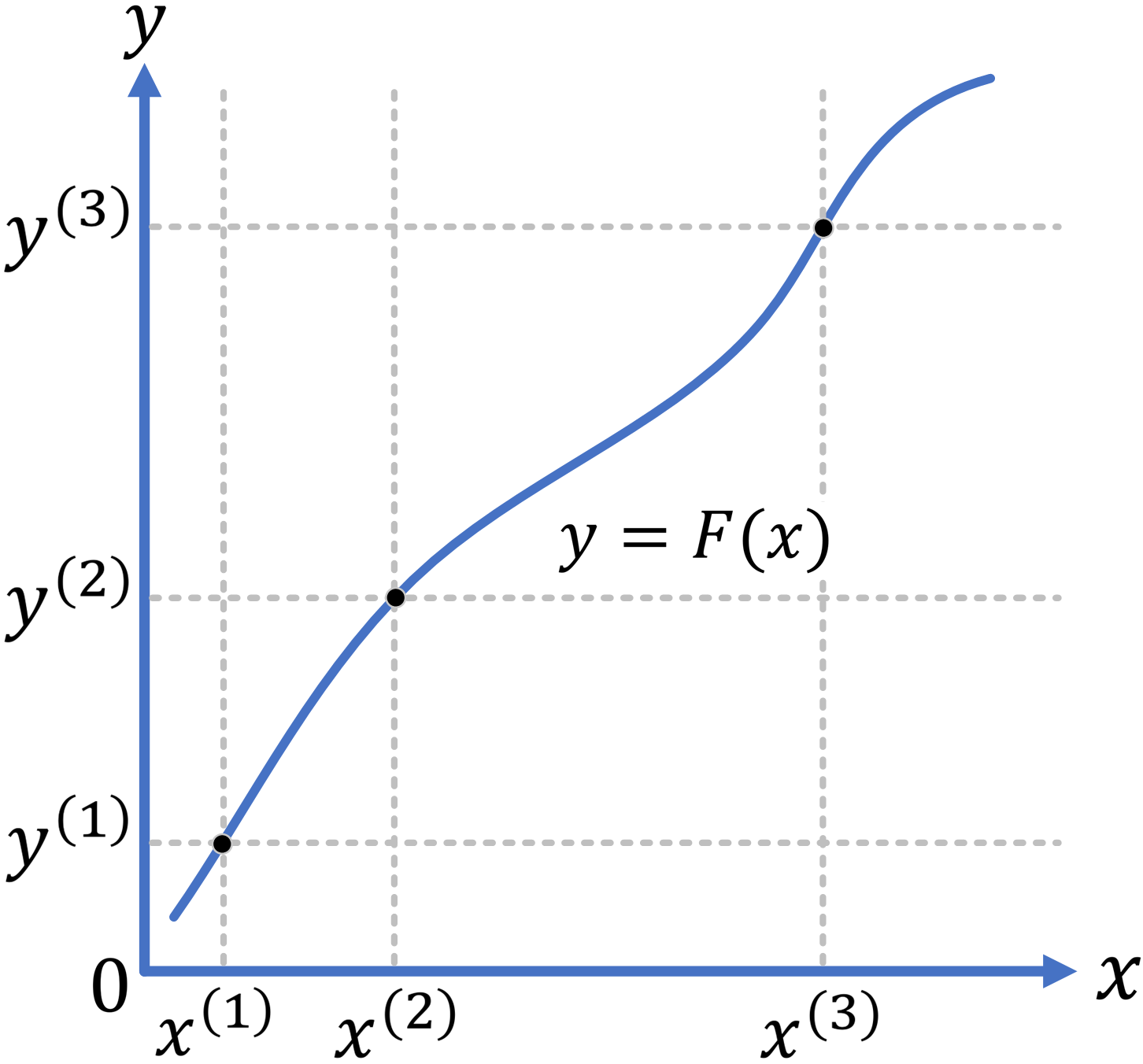}
	\centering
  	\caption{The fitting curve with a small number of samples.}
  	\label{fig:fig1}
\end{figure*}

\textbf{Situations where there are no same-type samples}\\
For the sake of visualization, in the discussion of this paragraph, we limit the sample size to 3. As shown in Figure \ref{fig:fig1}, let the dataset be $D=\{(x^{(i)},y^{(i)})|i \in [1,3]\}$, $(x^{(i)},y^{(i)})$ be the $i$-th sample, $x^{(i)}$ be the original representation of the sample, and  $y^{(i)}$  be the category to which  $x^{(i)}$  belongs. Our goal is to find a function $F$ for each  $x^{(i)}$  such that $y^{(i)}=F(x^{(i)})$. To reveal the true working principle of neural networks, we abandon the concepts of feature and label, and instead use \textbf{surface} and \textbf{essence} to refer to $x^{(i)}$ and $y^{(i)}$. Meanwhile, in order to grasp the key of the problem and simplify it, both the surface and the essence in Section \ref{sec: Ideal Model} are represented by scalars. The function $F$ shown in Figure \ref{fig:fig1} is the ideal model we hope to obtain, because for any surface $x^{(i)}$, it can precisely give the corresponding essence $y^{(i)}$.

\begin{figure*}[h]
  	\includegraphics[scale=1]{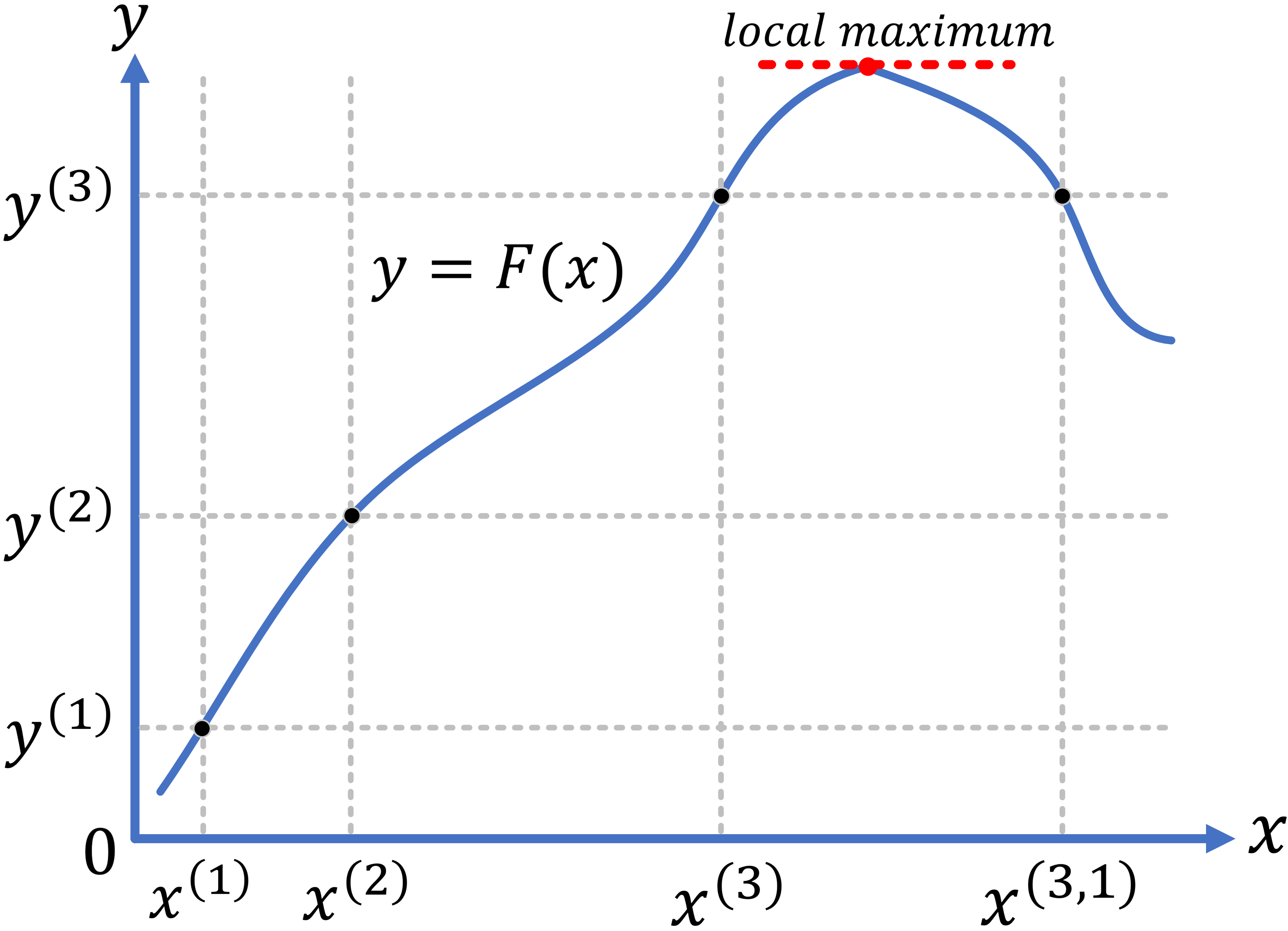}
	\centering
  	\caption{The fitting curve with a local maximum.}
  	\label{fig:fig2}
\end{figure*}

\noindent\textbf{Situations where there are same-type samples}\\
As shown in Figure \ref{fig:fig2}, if a new sample that is essentially the same as one in the dataset $D$ is added, for instance, adding sample $(x^{(3,1)},y^{(3)})$, then the function curve $F$ needs to change its shape so that the new sample can just fall onto the function curve. At this time, there is a local maximum between the samples $(x^{(3,1)},y^{(3)})$ and $(x^{(3)},y^{(3)})$ on the function curve.

\begin{figure*}[h]
  	\includegraphics[scale=1]{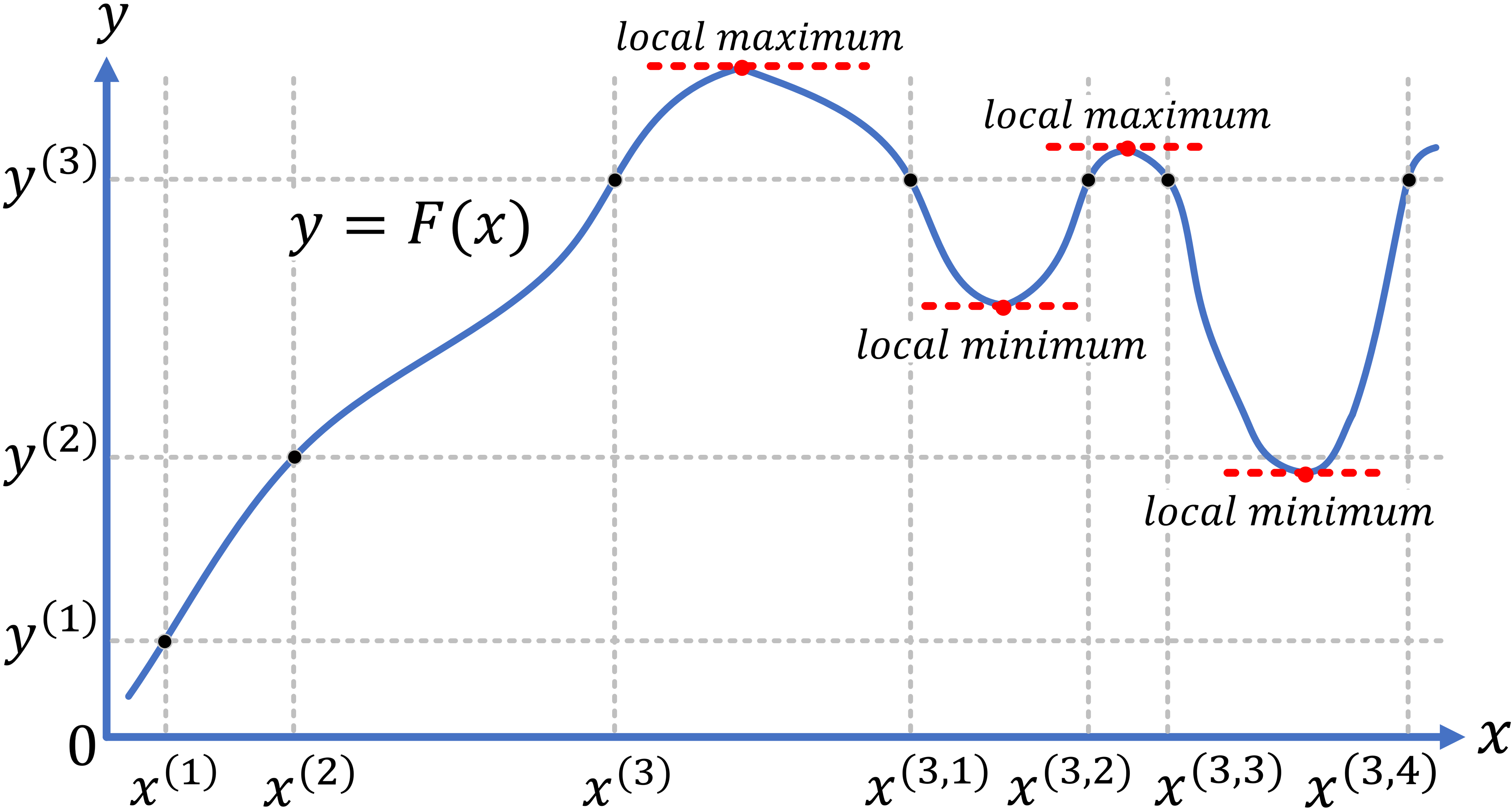}
	\centering
  	\caption{The fitting curve with more local maxima/minima.}
  	\label{fig:fig3}
\end{figure*}

Similarly, as shown in Figure \ref{fig:fig3}, if new samples with the essence of $y^{(3)}$, such as $(x^{(3,2)},y^{(3)})$, $(x^{(3,3)},y^{(3)})$, and $(x^{(3,4)},y^{(3)})$, are continuously added, the function curve needs to further change its shape to accommodate these new samples, thereby forming multiple local minima/maxima. If function $F$ can achieve such shape alteration, it possesses the ability to precisely map any surface to its essence, meaning it has true generalization capability.

\subsection{Weakened Mapping}
\begin{figure*}[h]
  	\includegraphics[scale=1.2]{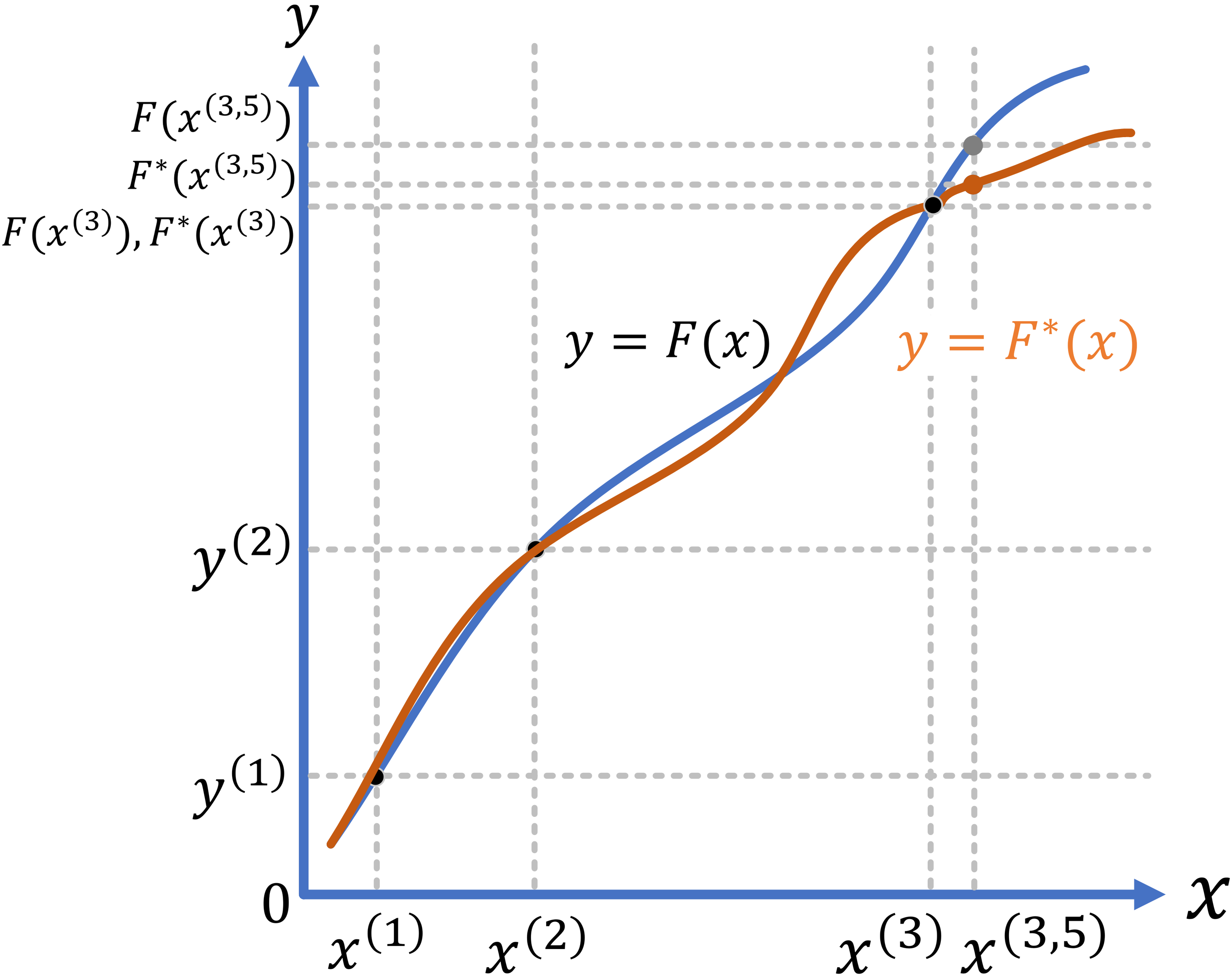}
	\centering
  	\caption{The approximately fitting curve with extended essence.}
  	\label{fig:fig4}
\end{figure*}

To obtain the aforementioned ideal function, the computation is usually enormous. For a function with a limited number of parameters, the degree of change in its curve shape is limited, and the number of extreme values cannot be increased at will. Then, how should we handle the situation when the surface of a sample has only changed slightly while its essence remains unchanged? A natural idea is to expand the essence from a single point to an interval, so that samples with slightly different surfaces but the same essence can be concentrated in this interval. As shown in Figure \ref{fig:fig4}, we add sample $(x^{(3,5)},y^{(3)})$ where the distance between $x^{(3,5)}$ and $x^{(3)}$ is small enough. We adjust the precise mapping function $F$ to the approximately fitting function $F^*$, making the difference between $F^* (x^{(3,5)})$ and $F^* (x^{(3)})$ as close as possible. When $|F^* (x^{(3,5)}) - F^* (x^{(3)})|$ is small enough, we can approximately consider that the surfaces falling within the interval $[F^* (x^{(3,5)}), F^* (x^{(3)})]$ all have the essence of $y^{(3)}$. Then we call function $F^*$ a weakened model of function $F$.

\noindent\textbf{Interval partition}\\
Each sample consists of both a surface and an essence. A surface is usually a one-dimensional vector or a multi-dimensional matrix. Once the algorithm for generating the surface is determined, for example, using a two-dimensional matrix to represent a grayscale image, where each matrix's element's value ranges from 0 to 255, then the surface is definite and we cannot make further changes to it. However, the essence is different. It is usually just an abstract concept and can be represented by any scalar or vector. As shown in Figure \ref{fig:fig4}, if the shape of the curve of the function $F$ is restricted, then each essence requires a tolerance interval. How is this interval selected? One method is to divide the range of the function $F$ into $N$ intervals with the same length, where $N$ is the total number of essence types. Then each interval is assigned to each essence, and the essences of surfaces that fall within the same interval are the same. When dividing essences using this method, the range of values of the objective function should be finite.

\subsection{N Classification to Binary Classification}
There is a problem in the interval partition method. When there are many types of essences and the value range of the function $F$ is limited within a small interval, for example, the value of each element in the output layer of a neural network is limited within a small interval $(0,1)$, then overlap is prone to occur. One solution is to reduce the types of essences and thereby expand each partition. But this introduces two new problems. One is to what extent the essence types should be reduced? Second, how should the essence of being excluded be handled?

For the above problems, we can reduce the number of essences to only one type. That is to say, the target model changes from an $N$-classification function $F$ to $N$ binary classification functions $\{F_j | j \in [1,N]\}$, and the $j$-th binary classification function $F_j$ only determines whether the input sample belongs to the $j$-th type of essences. That is, for any given sample $(x^{(i)},y^{(i)})$, where $i>0$, the ideal objective function $F_j$ satisfies:

\begin{equation*}
  F_j (x^{(i)}) =
    \begin{cases}
      $$UB$$, & \text{$y^{(i)}=j$}\\
      $$LB$$, & \text{$y^{(i)} \neq j$}
    \end{cases}       
\end{equation*}

The weakened objective function $F^*_j$ satisfies:

\[
  F_j ^*(x^{(i)}) \in
    \begin{dcases}
      $$(\dfrac{LB^*+UB^*}{2},UB^*]$$, & \text{$y^{(i)}=j$}\\
      $$[LB^*,\dfrac{LB^*+UB^*}{2})$$, & \text{$y^{(i)} \neq j$}
    \end{dcases}       
\]

\begin{figure*}[h]
  	\includegraphics[scale=1.8]{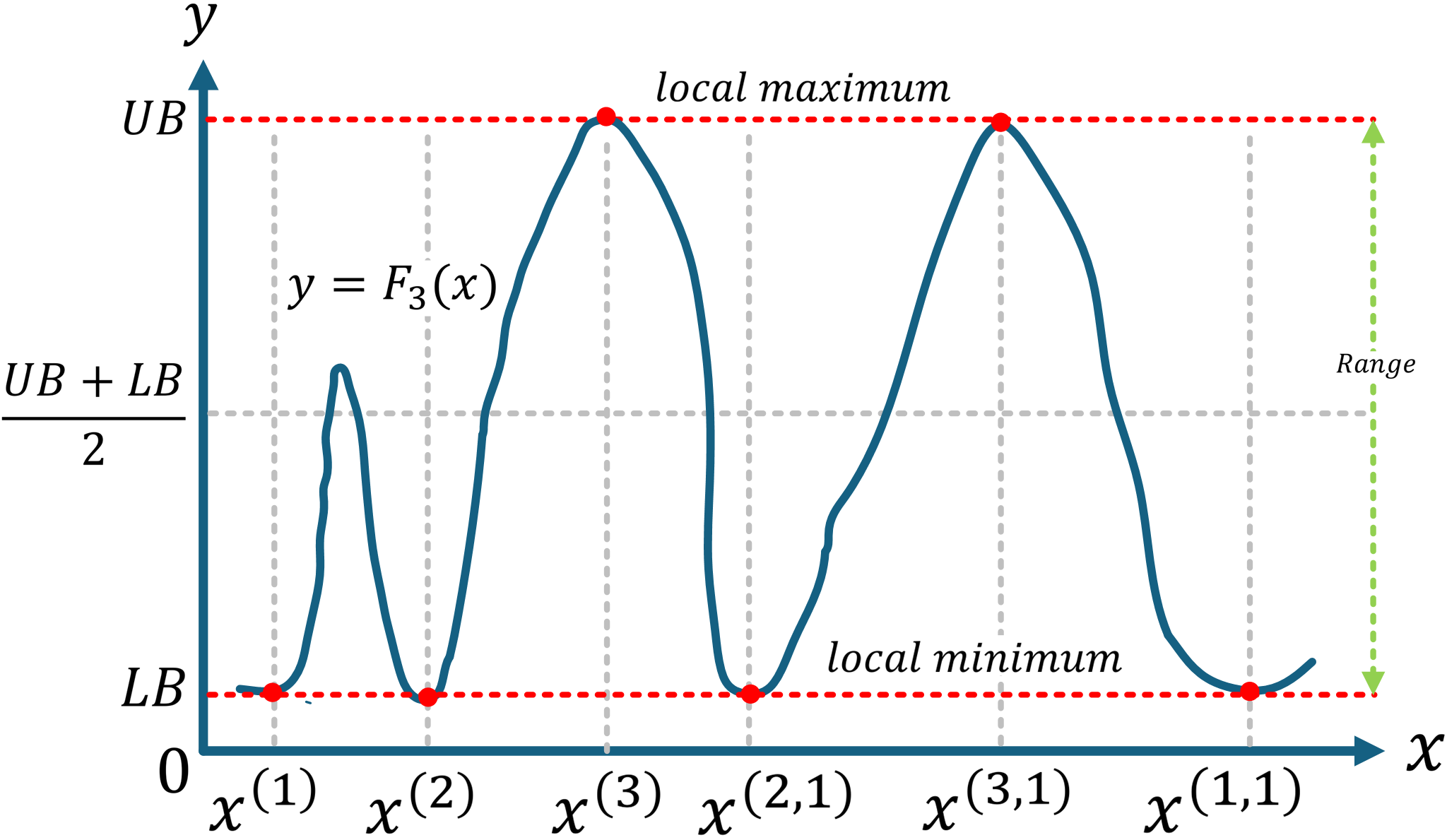}
	\centering
  	\caption{The ideal fitting curve for a binary classification problem.}
  	\label{fig:fig5}
\end{figure*}

$LB$ and $UB$ are the lower and upper limits of the function $F_j$ respectively, and correspondingly, $LB^*$ and $UB^*$ are the lower and upper limits of the function $F_j^*$. For the ideal function $F_j$, each given sample is adjusted to be its extremum point. Figure \ref{fig:fig5} presents an instance of a binary classification function $F_3$. We adjust the parameters of $F_3$ so that all samples of the third-essence are adjusted to the upper limit of the function's value range, all other-essence samples are adjusted to the lower limit. Correspondingly, the weakened function $F_3^*$ uses the midpoint of the value range as the dividing line. The same parameter adjustment is made for other binary classification functions (such as $F_1$, $F_2$).

\subsection{General Training Process of an Ideal Model}
In summary, the ideal training process for all machine learning models that are essentially classification problems can be summarized as the following steps:

\textbf{1)} Transform the $N$-class objective function $F$ into a family of binary classification functions $\{F_j | j \in [1, N]\}$ and initialize all parameters.

\textbf{2)} For each $F_j$, adjust the parameters so that each training surface $x^{(i)}$ is exactly one of the extrema of the function.

\textbf{3)} For each $F_j$, adjust the parameters so that the training samples of the $j$-th essence become a local maximum, and those of non-$j$-th essence become a local minimum.

\textbf{4)} Adjust the parameters to make the local maximum the global maximum and the local minimum the global minimum.

The model trained through the above steps can accurately map the input to the output, thereby enabling the machine to precisely answer ``yes" or ``no". In Section \ref{sec: neural network}, it was demonstrated that neural networks can be decomposed into a set of binary classification functions, with the corresponding surfaces mapped to the local extrema of these functions by finding the general solution of a homogeneous system of linear equations, thereby establishing the validity of Steps 1 and 2. In Section \ref{sec: EI algorithm}, a method for mapping the surfaces to the global extrema of the functions by enumerating the particular solutions of the homogeneous system of linear equations was presented, thus confirming Steps 3 and 4.

\section{Situations on Neural Networks}\label{sec: neural network}
\subsection{Model Decomposition}
Any type of neural network, whether it is a traditional artificial neural network, an improved convolutional neural network, or a recurrent neural network, is composed of three parts: an input vector with a fixed number of elements, an intermediate processing layer with undetermined parameters, and an output vector with the number of elements equal to the number of  essence types. To reduce computational complexity and focus on the main working process of neural networks, we only conduct derivative analysis on the fully connected neural networks. Additionally, the output layer of mainstream neural networks often uses the softmax function, which is just a normalization operation added to the sigmoid function. To simplify the operation steps, we directly use the sigmoid function as the output layer, so that both the hidden layers and the output layer use the sigmoid function. Moreover, we remove the biases because they are irrelevant to the essential attributes of the model but make the calculations lengthy and reduce readability. Then on this basis, we analyze the structure of a fully connected neural network based on an $N$-classification problem.

\begin{figure*}[h]
  	\includegraphics[scale=1.8]{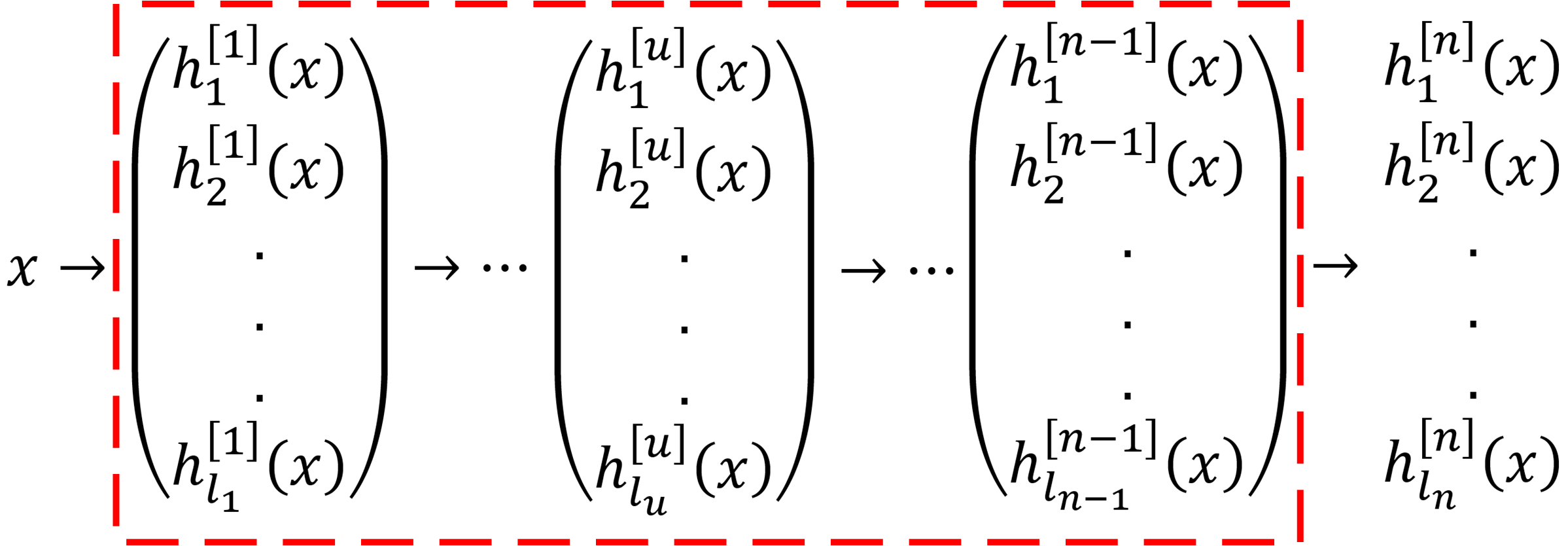}
	\centering
  	\caption{A simplified neural network topology.}
  	\label{fig:fig6}
\end{figure*}

Figure \ref{fig:fig6} is a schematic diagram of a fully connected neural network structure expressed directly through numerical relations and without graphical representation. Each sample is denoted as $(x, y)$, where the surface $x$ is an $m$-dimensional column vector, $x = (x_1, x_2, …, x_m)^T$, and $y \in [1, l_n]$ is the essence corresponding to $x$, where $l_n$ is the number of elements in the neural network's output vector, representing the number of essence types. The neural network has a total of $n$ layers with the same processing method, with the first $n - 1$ layers being hidden layers and the $n$-th layer being the output layer. The total number of elements in the $u$-th layer (with the input vector being the $0$-th layer) is denoted as $l_u$, and the $v$-th element in the $u$-th layer is denoted as $h_v^{[u]} (x)$, where $v \in [1, l_u]$. As can be seen from Figure \ref{fig:fig6}, disregarding the dazzling connection of the neurons, a neural network is actually a set composed of $l_n$ composite functions $\{h_v^{[n]} (x) | v \in [1, l_n]\}$, with each composite function sharing the same hidden layers.

\begin{figure*}[h]
  	\includegraphics[scale=1.8]{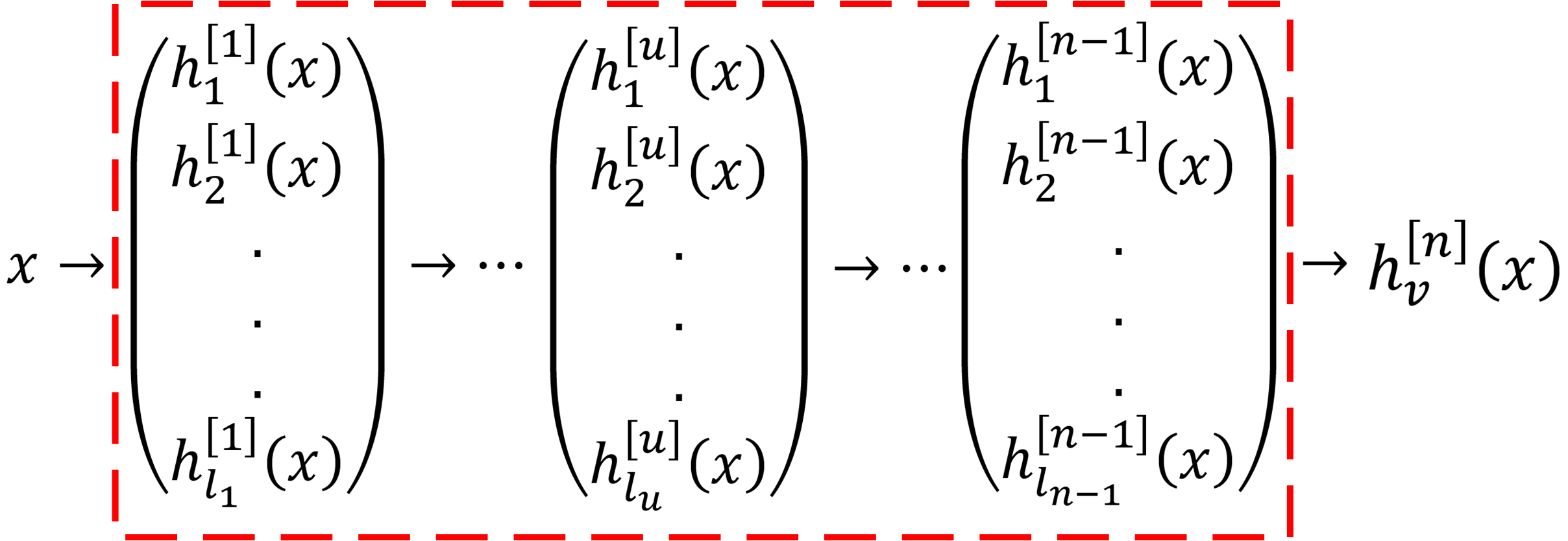}
	\centering
  	\caption{The basic component of neural networks is a binary classification.}
  	\label{fig:fig7}
\end{figure*}

For samples belonging to the $v$-th essence where $v \in [1, l_n]$, the target output vector of the neural network is $(0,... ,h_v^{[n]}(x)=\textbf{1},…,0)^T$. For all other essence types, the target output vector of the neural network is $(\omega,...,h_v^{[n]}(x)=\textbf{0},…,\omega)^T$ where one of the $\omega$ is 1 and the others are all 0.  In this simplified model, It is worth mentioning that when the sigmoid function is used as the output, the upper and lower limits of $h_v^{[n]}(x)$ can only approach 1 and 0 infinitely. When we transform the output layer that seems to be composed of an $l_n$-dimensional vector into $l_n$ scalars, the entire model becomes very clear. That is, each composite function $h_v^{[n]}(x)$ is actually a binary classification problem of the $v$-th essence. Therefore, a neural network with a multi-dimensional vector output layer can actually be regarded as a collection of multiple binary classification functions as shown in Figure  \ref{fig:fig7}. We can analyze each function $h_v^{[n]}(x)$ separately and then integrate them to obtain the characteristics of the neural network.

\subsection{Extreme Points of the Model}
Specifically, the expression of the function $h_v^{[u]}(x)$ satisfies:

\begin{equation*}
    \begin{dcases}
      $$h_v^{[u]}(x)=S(\sum\limits_{k=1}^{l_{u-1}} w_{v,k}^{[u]}*h_k^{[u-1]}(x))$$, & \text{$u>1$}\\
      $$h_v^{[1]}(x)=S(\sum\limits_{k=1}^m w_{v,k}^{[1]}*x_k)$$, & \text{$u=1$}
    \end{dcases}      
 \end{equation*}

The function $S(\theta) = \dfrac{1} {1 + e^{-\theta}}$ is the sigmoid function, and $w_{v,k}^{[u]}$ represents the parameters between the $(u - 1)$-th layer and the $u$-th layer of the neural network, which is the same as the parameters in traditional neural networks. To further enhance readability, let $z_v^{[u]}(x) = \sum\limits_{k=1}^{l_{u-1}} w_{v,k}^{[u]}*h_k^{[u-1]}(x)$. Then we take the partial derivative of the function $h_v^{[u]}(x)$:

\[
\frac{\partial}{\partial x_t} h_v^{[u]}(x) = \frac{\partial}{\partial x_t} S(z_v^{[u]}(x))  
\]
\[
=  \frac{\partial}{\partial z_v^{[u]}(x)} S(z_v^{[u]}(x)) *\frac{\partial}{\partial x_t} z_v^{[u]}(x)
\]
\[
= S(z_v^{[u]}(x))*(1-S(z_v^{[u]}(x)))*\frac{\partial}{\partial x_t} z_v^{[u]}(x)
\]

where $t \in [1,m]$, and the derivative result of the function $S(\theta)$ is adopted, i.e., $\frac{\partial}{\partial \theta} S(\theta)=S(\theta)*(1-S(\theta))$. Let  $c_v^{[u]}(x)=S(z_v^{[u]}(x))*(1-S(z_v^{[u]}(x)))$, then

\[
\frac{\partial}{\partial x_t} h_v^{[u]}(x) = c_v^{[u]}(x)*\frac{\partial}{\partial x_t} z_v^{[u]}(x) 
\]
\[
=  c_v^{[u]}(x)*\frac{\partial}{\partial x_t} \sum\limits_{k=1}^{l_{u-1}} w_{v,k}^{[u]}*h_k^{[u-1]}(x) 
\]
\[
= c_v^{[u]}(x)*\sum\limits_{k=1}^{l_{u-1}} w_{v,k}^{[u]}* \frac{\partial}{\partial x_t} h_k^{[u-1]}(x) 
\]

Let $\frac{\partial}{\partial x_t} h_v^{[u]}(x) = 0$, since $c_v^{[u]}(x)>0$, then $\sum\limits_{k=1}^{l_{u-1}} w_{v,k}^{[u]}* \frac{\partial}{\partial x_t} h_k^{[u-1]}(x)=0$. Then starting from the output layer, i.e., $u = n$, we take the partial derivatives of all components of $x$ respectively, and obtain the following system of equations:

\[
L(n,v) =
	\begin{dcases}
	\begin{array}{c @{\thickspace}c @{\thickspace}c l}
		$$\sum\limits_{k=1}^{l_{n-1}} w_{v,k}^{[n]}* \frac{\partial}{\partial x_1} h_k^{[n-1]}(x)=0$$\\
      		$$\sum\limits_{k=1}^{l_{n-1}} w_{v,k}^{[n]}* \frac{\partial}{\partial x_2} h_k^{[n-1]}(x)=0$$\\
	      	\vdots\\
      		$$\sum\limits_{k=1}^{l_{n-1}} w_{v,k}^{[n]}* \frac{\partial}{\partial x_m} h_k^{[n-1]}(x)=0$$
	\end{array} 
    	\end{dcases}    
\]

When a surface $x$ is given, the above system of equations is a homogeneous linear equation system consisting of $m$ equations, $l_{n-1}$ independent variables $\{w_{v,k}^{[n]} | k \in [1, l_{n-1}]\}$ and $m*l_{n-1}$ coefficients $\{\frac{\partial}{\partial x_t} h_k^{[n-1]}(x) | k \in [1, l_{n-1}], t \in [1, m]\}$. Let the rank of the coefficient matrix of $L(n, v)$ be $r(n, v)$, then when $r(n, v)$ is less than the number of unknowns $l_{n-1}$, the linear equation system has infinitely many solutions. Since $r(n, v) \leq m$, as long as the number of neurons in the last hidden layer $l_{n-1}$ is greater than $m$ when designing a neural network, we can always find infinitely many parameter combinations that make the surface $x$ be an extremum point of the binary classification function $h_v^{[n]} (x)$ for $v \in [1, l_n]$. This is the main reason why neural networks have strong generalization ability, and the black box begins to be unveiled.

The shapes of the curves for the other binary classification functions can be adjusted simultaneously. Let 

\[
L(n,-) =
	\begin{dcases}
	\begin{array}{c @{\thickspace}c @{\thickspace}c l}
		$$L(n,1)$$\\
      		$$L(n,2)$$\\
	      	\vdots\\
      		$$L(n,l_n)$$
	\end{array} 
    	\end{dcases}    
\]

When given a surface $x$, $L(n,-)$ is a homogeneous linear system of equations consisting of $m*l_n$ equations, $l_n*l_{n-1}$ variables $\{w_{v,k}^{[n]} | v \in [1,l_n], k \in [1,l_{n-1}]\}$, and $m*l_{n-1}$ coefficients $\{\frac{\partial}{\partial x_t} h_k^{[n-1]}(x) | k \in [1, l_{n-1}], t \in [1, m]\}$. Any solution of $L(n,-)$ makes the surface $x$ be the extremum point of each binary classification function. Then we can select a particular solution such that when the surface $x$ belongs to the $v$-th essence, the corresponding extremum is the maximum value, and when $x$ belongs to other essences, the extremum is the minimum value. That is, $h_v^{n} (x)$ satisfies the ideal termination condition of model training:

\begin{equation}
\label{eq:ideal termination condition}
	h_v^{n} (x) =
	\begin{cases}
		$$1$$, & \text{$y=v$}\\
		$$0$$, & \text{$y \neq v$}
	\end{cases} 
	\text{, \hspace{2mm} $y \in [1,l_n]$}	      
\end{equation}

If it is difficult to find the above particular solution, then the constraints can be relaxed, that is, a weakened termination condition can be adopted:

\begin{equation}
\label{eq:weakened termination condition}
	h_v^{[n]} (x) \in
	\begin{cases}
		$$(0.5,1]$$, & \text{$y=v$}\\
		$$[0,0.5)$$, & \text{$y \neq v$}
	\end{cases} 
	\text{, \hspace{2mm} $y \in [1,l_n]$}	      
\end{equation}

\subsection{Continuous Optimization of Parameter Combinations}
The above discussion only covers the situation where there is only one training sample. What should be done when the number of samples increases? Let 

\[
L(n,v,x^{(i)}) =
	\begin{dcases}
	\begin{array}{c @{\thickspace}c @{\thickspace}c l}
		$$\sum\limits_{k=1}^{l_{n-1}} w_{v,k}^{[n]}* \frac{\partial}{\partial x_1} h_k^{[n-1]}(x)\vert_{x=x^{(i)}}=0$$\\
      		$$\sum\limits_{k=1}^{l_{n-1}} w_{v,k}^{[n]}* \frac{\partial}{\partial x_2} h_k^{[n-1]}(x)\vert_{x=x^{(i)}}=0$$\\
	      	\vdots\\
      		$$\sum\limits_{k=1}^{l_{n-1}} w_{v,k}^{[n]}* \frac{\partial}{\partial x_m} h_k^{[n-1]}(x)\vert_{x=x^{(i)}}=0$$
	\end{array} 
    	\end{dcases}    
\]

Then, 

\[
L(n,-,x^{(i)}) =
	\begin{dcases}
	\begin{array}{c @{\thickspace}c @{\thickspace}c l}
		$$L(n,1,x^{(i)})$$\\
      		$$L(n,2,x^{(i)})$$\\
	      	\vdots\\
      		$$L(n,l_n,x^{(i)})$$
	\end{array} 
    	\end{dcases}    
\]

When we train the neural network with a dataset $\Phi = \{(x^{(i)}, y^{(i)}) | i \in [1, \phi]\}$, we are actually solving the following homogeneous linear equation system:

\[
L(n,-,\Phi) =
	\begin{dcases}
	\begin{array}{c @{\thickspace}c @{\thickspace}c l}
		$$L(n,-,x^{(1)})$$\\
      		$$L(n,-,x^{(2)})$$\\
	      	\vdots\\
      		$$L(n,-,x^{(\phi)})$$
	\end{array} 
    	\end{dcases}    
\]

If $L(n, -, \Phi)$ has infinitely many solutions, then a particular solution that meets the conditions can be found from the general solution of the system of equations. Otherwise, parameters between the $(n-2)$-th layer and the $(n-1)$-th layer need to be introduced. That is, we need to expand the partial derivatives in the system of equations $L(n,v)$ again. We substitute $\frac{\partial}{\partial x_t} h_k^{[n-1]}(x) = c_k^{[n-1]}(x)*\sum\limits_{p=1}^{l_{n-2}} w_{k,p}^{[n-1]}* \frac{\partial}{\partial x_t} h_p^{[n-2]}(x)$ into $L(n,v)$, and simplify the system:

\[
L(n-1,v) =
	\begin{dcases}
	\begin{array}{c @{\thickspace}c @{\thickspace}c l}
		$$\sum\limits_{k=1}^{l_{n-1}} w_{v,k}^{[n]}* \sum\limits_{p=1}^{l_{n-2}} w_{k,p}^{[n-1]}* \frac{\partial}{\partial x_1} h_p^{[n-2]}(x)=0$$\\
      		$$\sum\limits_{k=1}^{l_{n-1}} w_{v,k}^{[n]}* \sum\limits_{p=1}^{l_{n-2}} w_{k,p}^{[n-1]}* \frac{\partial}{\partial x_2} h_p^{[n-2]}(x)=0$$\\
	      	\vdots\\
      		$$\sum\limits_{k=1}^{l_{n-1}} w_{v,k}^{[n]}* \sum\limits_{p=1}^{l_{n-2}} w_{k,p}^{[n-1]}* \frac{\partial}{\partial x_m} h_p^{[n-2]}(x)=0$$
	\end{array} 
    	\end{dcases}    
\]

Similarly, we can obtain the system of equations $L(n-1,-)$, which can be regarded as a homogeneous nonlinear system of equations consisting of $m*l_n$ equations with $l_{n-1}*l_n+l_{n-2}*l_{n-1}$ independent variables $\{w_{v,k}^{[n]} | v \in [1,l_n], k \in [1,l_{n-1}]\}$ and $\{w_{k,p}^{[n-1]} | k \in [1,l_{n-1}], p \in [1,l_{n-2}]\}$. Although $L(n-1,-)$ seems to be a nonlinear system of equations, due to its very regular structure, for instance, $w_{1,1}^{[n]} * w_{1,1}^{[n-1]}$ can be regarded as a whole, then the solution method for homogeneous linear equations can still be adopted. Then we just need to find the particular solution of the equation system $L(n-1,-,\Phi)$ that meets the requirements. By solving the homogeneous equations layer by layer, the dataset can be mapped to the neural network.

\section{The EI Algorithm}\label{sec: EI algorithm}
\subsection{General Training Method}
From the above discussion, we have obtained a preliminary model training framework, which we call the EI algorithm. Its main steps have significant differences from the current commonly used neural network training methods, such as the BP algorithm. Firstly, the BP algorithm uses gradient updates to approximate the ideal values of parameters, while the EI algorithm attempts to directly obtain the values of parameters by solving systems of equations. Secondly, the BP algorithm needs to update all parameters each time, while the EI algorithm only needs to update some parameters. Debugging all training samples to the extremum points of the model is the key to the entire framework. In this subsection, we will have a more in-depth discussion on the details of the algorithm.

Table \ref{tb:1} shows the state of neural network parameters based on EI algorithm at each round where $W[u] = \{w_{v,k}^{[u]} | v \in [1, l_u], k \in [1, l_{u-1}]\}$, $init$ indicates that the parameters remain at their initial values, and $update$ indicates that the parameters are updated in the current round. In the first round, we first solve the equation system $L(n, -, \Phi)$. If there is a solution, only the parameters $W[n]$ need to be updated. Otherwise, in the second round, we solve the equation system $L(n-1, -, \Phi)$. If there is a solution, the parameters $W[n]$ and $W[n-1]$ need to be updated. Otherwise, we solve the equation system $L(n-2, -, \Phi)$, and so on.

\begin{table}
\caption{Weights’ states in different stages.}
\label{tb:1}
\[
\begin{array}{@{}l*{6}{c}@{}}
\toprule
\text{stage} & \multicolumn{6}{c@{}}{\text{weight}}\\
    \cmidrule(l){2-7}
    & W[1] & W[2] & \ldots & W[n-2] & W[n-1]& W[n]\\
\midrule
L(n,-,\Phi) & init & init & \ldots & init & init& update\\
L(n-1,-,\Phi) & init & init & \ldots & init & update& update\\
L(n-2,-,\Phi) & init & init & \ldots & update & update& update\\
\ldots & \ldots & \ldots & \ldots & \ldots & \ldots & \ldots\\
\bottomrule
\end{array}
\]
\end{table}

Algorithm \ref{alg:1} presents the main steps for precisely mapping a dataset to the neural network model. The symbols used, unless otherwise specified, have the same meanings as those in the previous text. In the initial stage of the algorithm, we manually label the sample set $\Phi$. If a sample $(x^{(i)}, y^{(i)})$ is classified as the $j$-th essence where $j \in [1, l_n]$, then $(x^{(i)}, y^{(i)}) = (x^{(i)}, j)$. After that, we initialize the parameter set $W$ to non-zero real numbers.

\begin{algorithm}
	\label{alg:1}
        \caption{Precise mapping from input to output}
        \begin{algorithmic}[1] 
            \Require $\phi=\{(x^{(i)},y^{(i)}) | i \in [1,\varphi]\}$
            \Ensure $W=\{w_{v,k}^{[u]} | u \in [1,n],v \in [1,l_u],k \in [1,l_{u-1}]\}$
            \Function {FittingCurve}{$ $}
             	\State \Call{Init}{$W$}
            	\For {$u \in [1,n-1],v \in [1,l_u],t \in [1,m],i \in [1,\varphi]$}			
			\State \Call {Calculate} {$\left.\frac{\partial}{\partial x_t} h_{v}^{[u]}(x)\right|_{x=x^{(i)}} $} \Comment{for calculating $W{[u:n]}$}
		\EndFor
		\For {$u \in [1,n-1],v \in [1,l_u],i \in [1,\varphi]$}			
			\State \Call {Calculate} {$h_{v}^{[u]}(x^{(i)})$} \Comment{for calculating $W^{[u:n]}$}
		\EndFor
		
		\State  $u \gets n$
		\While{$u \geq 1$}
			\State $W[u:n] \gets$ \Call {$L$}{$u,-,\phi$} \Comment{the general solution, equals to $\{W[j]|j \in [u,n]\}$}
			\State $ W^{[u:n]} \gets $ \Call {Polarize} {$\{h^{[n]}_{v}(x)| v \in [1,l_n]\},W[u:n],\phi$} \Comment{the particular solution}
				 
			\If {$W^{[u:n]} \neq \{\} \land W^{[u:n]} \neq \{0\}$}
				\State $W \gets$ \Call {Update}{$W^{[u:n]}$}; $break$
			\Else
				\State $ u--$
			\EndIf
		\EndWhile
				
		\If {$u \geq 1$}
			\Return $W$
		\Else
			\ \Return \Call {Error}{$ $}
		\EndIf
            \EndFunction
        \end{algorithmic}
    \end{algorithm}

Just like the BP algorithm, the parameter update is executed layer by layer from the last hidden layer to the first hidden layer. We first calculate the values and partial derivatives of all the neurons in the hidden layers, and then solve the general solution $W[u:n] = \{W[j] | j \in [u, n]\}$ of the equation group $L(u, -, \Phi)$. Then, we select a particular solution $W^{[u:n]}$ that satisfies the termination condition \ref{eq:ideal termination condition} from $W[u:n$]. We call this operation \textbf{polarize}. If a particular solution $W^{[u:n]}$ is found, the parameters $W$ are then updated. If no particular solution is found, it indicates that the parameter set $W[u:n]$ cannot precisely map the sample set $\Phi$ to the neural network. Then, we need to introduce the parameter $W[u-1]$ to find the particular solution $W^{[u-1:n]}$ of the equation group $L(u-1, -, \Phi)$. If no particular solution that meets the requirements is found after traversing all the parameters of the neural network, we need to consider adjusting the structure of the neural network (such as increasing the number of hidden layers or the number of nodes in each hidden layer).

\subsection{Reduce the Computational Complexity}\label{subsec: softmax}
In Algorithm \ref{alg:1}, the polarization time for selecting a specific solution $W^{[u:n]}$ from the general solution $W[u:n]$ is uncertain. This is because we do not know the characteristics of the specific solution and can only verify each instance of the general solution through enumeration. To reduce the training time, we can relax the termination condition of the model training. That is, when a sample $(x^{(i)}, y^{(i)})$ is a sample of the $v$-th essence, the value of the $v$-th binary classification function $h_v^{[n]} (x^{(i)})$ just need to be much greater than the values of any other binary classification function $h_q^{[n]} (x^{(i)})$, without considering whether these values are maximum or minimum values. This is equivalent to the following weakened condition:

\[
	\begin{cases}
		$$1-\frac{h_v^{[n]}(x^{(i)})} {\sum_{j=1}^{l_n} h_j^{[n]}(x^{(i)})} < \alpha$$, & \text{$v \in [1,l_n]$}\\
		$$\frac{h_q^{[n]}(x^{(i)})} {\sum_{j=1}^{l_n} h_j^{[n]}(x^{(i)})} - 0 < \beta$$, & \text{$any \, q \in [1,l_n], q \neq v$}
	\end{cases} 
	\text{, \hspace{2mm} $y^{(i)} = v$}
\]	      

where $\alpha$ and $\beta$ are two sufficiently small positive real numbers. This is the situation when the softmax function is used as the output layer of the neural network. That is to say, a neural network using the softmax function can be regarded as a weakened version of an ideal model.

\subsection{Reduce the Computational Scale}\label{subsec: reduce scale}
If each training sample corresponds to an extreme point on the model curve, that is, by adding an equation set $L(n, -, x^{(i)})$, the required scale of network parameters will be extremely large and the training time will also increase significantly. Is there a way to reduce the number of equation sets? To this end, we propose the concept of surface \textbf{neighborhood}. In a further weakened neural network, only a portion of the samples need to be the extreme points, and the other samples can only satisfy the weakened termination condition \ref{eq:weakened termination condition}. Then which samples can have their restrictions relaxed? An intuitive idea is that only the representative of all adjacent samples needs to satisfy the strict condition. Let $A = (x^{(a)}, y^{(a)})$ and $B = (x^{(b)}, y^{(b)})$ be two samples in the  dataset $\Phi= \{(x^{(i)}, y^{(i)}) | i \in [1, \phi]\}$ where $a,b \in [1,\phi]$. Then the \textbf{distance} between these two samples is defined as:

\[
	Ds(A,B)=\sqrt[2]{\sum\limits_{j=1}^{dim(x)}(x_j^{(a)}-x_j^{(b)})^2}
\]	      

If $A$ and $B$ are samples of the same essence, that is, $y^{(a)} = y^{(b)}$, and the proximity criterion is satisfied:

\[
	Ds(A,B) < \gamma
\]	      

where $dim(x)$ represents the dimension of a sample surface, and $\gamma$ is a sufficiently small positive real number. Then we say that samples $A$ and $B$ of the same essence are located within each other's neighborhood. Due to the continuity of the function, it can be known that the function values of samples $A$ and $B$ are close on each binary classification function. Thus, one of the samples does not need to be sent to the algorithm for training but only needs to verify its function value. Then a further weakened training algorithm can be adjusted as follows:

\textbf{1)} Manually classify the training sample set and designate it as the major category.

\textbf{2)} Utilize a certain numerical algorithm, such as clustering algorithm, to further divide the samples of each major category into several minor categories. Each minor category has a central sample.

\textbf{3)} Train the model for all the central samples.

\textbf{4)} After the training is completed, verify whether the predicted values of all non-central samples on the neural network are within the specified accuracy range. If the accuracy requirements are met, the algorithm ends. Otherwise, mark the non-central samples that do not meet the requirements as central samples and repeat steps 3 and 4.

\section{Deductions}\label{sec: deductions}
\subsection{Gradient Vanishing/Explosion}
The problem of gradient vanishing/exploding is a common and very difficult issue encountered during the training of neural networks, and it is particularly prone to occur in deep networks. To address this problem, some scholars have proposed various methods to alleviate the adverse effects of gradient vanishing/exploding on parameter updates, such as using batch normalization \citep{santurkar2018does} and LSTM architecture \citep{yu2019review}. In the BP algorithm, the problem of gradient vanishing/exploding is often regarded as an abnormal issue that should be avoided.

Regarding the problem of gradient vanishing, as discussed in Sections \ref{sec: neural network} and \ref{sec: EI algorithm}, after the initialization of network parameters, the number of parameter updates required by the neural network varies depending on the sample size. If the particular solution $W^{[u:n]}$ can be found from the  general solution $W[u:n]$, then the parameters $W[1:u-1]$ of the earlier hidden layers can remain at their initial values. That is to say, according to the neural network characteristics revealed by the EI algorithm, gradient vanishing is an inevitable result.

Gradient explosion is also a similar problem. In the EI algorithm, when we calculate the equation set $L(1,-,\Phi)$, there may be cases where no solution exists. That is to say, the value of the solution is infinity, which corresponds to the gradient explosion in the BP algorithm. If the EI algorithm is adopted, simply increasing the number of hidden layers or the number of parameters in each layer will suffice.

\subsection{Overfitting}
The overfitting problem \citep{santos2022avoiding} seems different from the gradient vanishing/explosion problem, but in essence, both are caused by the same operational process. In the EI algorithm, if the equation set $L(1, -, \Phi)$ has solutions, but when we increase the number of the samples, the equation set $L(1, -, \Phi^\Delta)$ maybe has no solution where $\Phi \subset \Phi^\Delta$. That is, the neural network with the current parameter scale can only accommodate a limited number of samples $\Phi$, manifested as the overfitting phenomenon of the BP algorithm. This is an inherent characteristic of neural networks that there are only a limited number of extreme values under the condition of  limited parameters. Rather than saying it's overfitting, we would rather say it fits just right.

The BP algorithm reduces the model's dependence on the trained samples by adding noise to the samples and network parameters \citep{ying2019overview}. This method is similar to the clustering operation described in Section \ref{subsec: reduce scale}, which enables a fixed-structure neural network to accommodate more samples, but this often comes at the cost of model accuracy. Another approach is to increase the number of hidden layers or the number of parameters in each layer, that is, to increase the number of independent variables in the equation set $L(1,-,\Phi^\Delta)$, thereby accommodating more samples without sacrificing accuracy, at the cost of an increased training time.

\subsection{Adding noise}
During the training process of neural networks, we often enhance their robustness by adding noise to the existing samples and re-feeding them into the neural network for training \citep{xia2022gan}. This is because we have observed that after adding noise, even when humans do not perceive much difference between the previous and subsequent samples, the prediction accuracy of the machine drops sharply. This phenomenon can be explained by the concept of neighborhood. Let the initial sample be $A=(x,y)$ where $x=(x_1,x_2,\dots, x_m)^T$, and the noisy sample be $A^\Delta=(x^\Delta,y^\Delta)$ where  $x^\Delta=(x_1+x_1^\Delta, x_2+x_2^\Delta,\dots, x_m+x_m^\Delta)^T$, then

\[
	Ds(A,A^\Delta)=\sqrt[2]{\sum\limits_{j=1}^m(x_j^{\Delta})^2}
\]

The noisy sample may significantly deviate from the neighborhood of the original sample. If it is not within the neighborhood of other same-essence samples either, then the neural network will be unable to correctly process this sample, that is, $y^\Delta \neq y$. If there are too many noisy samples, it is difficult for the model to converge because we can add random noise. This is why we call the input vector of a neural network a ``surface", as there is a significant difference between what a neural network perceives and what humans see.

\subsection{Shallow/Deep Networks}
From the discussions in Sections \ref{sec: neural network} and \ref{sec: EI algorithm}, it can be concluded that the number of samples that a neural network can precisely fit is mainly positively correlated with the total number of network parameters, and has no necessary relationship with the depth of the network structure. If the number of samples is limited, we can directly adopt a network structure with only one hidden layer. According to the condition that homogeneous linear equations have a general solution, the number of parameters of a single-hidden-layer network should be greater than the product of the sample number, the surface dimension, and the essence types. If the number of samples is large and can be dynamically increased, we can adopt a ``tilted trapezoidal" network structure, in which the parameters of the last hidden layer are the most, and then the number of parameters decreases successively towards the first hidden layer. That is, in the EI algorithm, the calculation of invalid equation sets is minimized as much as possible.

\subsection{Probability}
\begin{figure*}[h]
  	\includegraphics[scale=1.4]{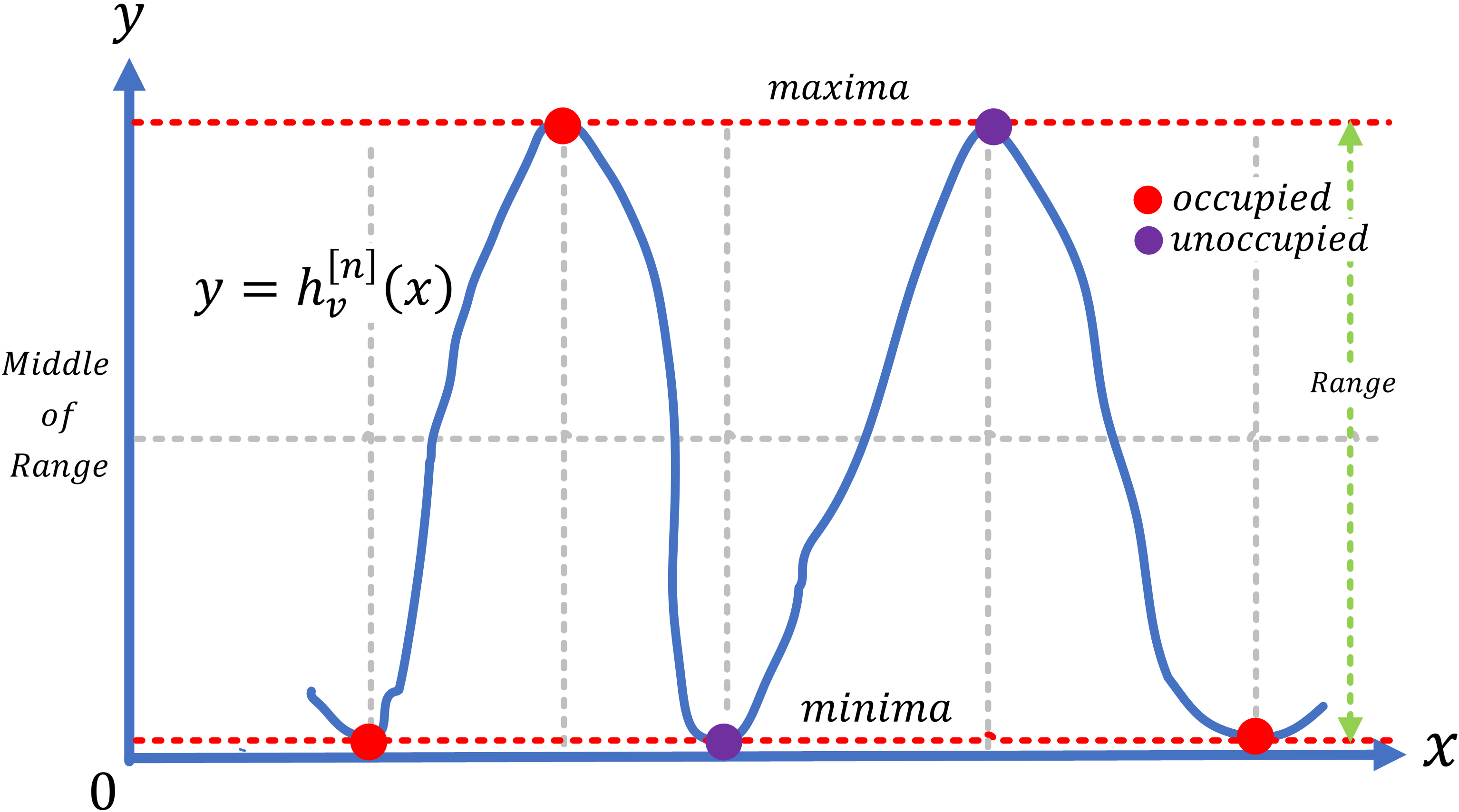}
	\centering
  	\caption{The extremum points of undetermined essences on $h_v^{[n]} (x)$.}
  	\label{fig:fig8}
\end{figure*}

The traditional view holds that the output layer of a neural network provides the probability that a surface of the input layer belongs to different essences. We believe this view is not entirely accurate, at least not in the strictly statistical sense of probability. In statistics, the probability of a random event is defined as the ratio of the certain output to the total output. No matter how large our training sample set is, we cannot exhaust or nearly exhaust the entire sample space, and there is no clear specific relationship between the finite sample set and the infinite sample space. For instance, we can add various noises to the existing samples, and the sample set can be easily expanded several times or even infinitely. Additionally, as shown in Figure \ref{fig:fig8}, the training sample set does not necessarily occupy all the extreme points of the trained binary classification function $h_v^{[n]} (x)$. Those unoccupied maximum points are not necessarily occupied by the $v$-th essence samples, and the minimum points are not necessarily occupied by non-$v$-th essence samples. In extreme cases, even if there is a sample that makes $h_v^{[n]} (x) = 1$ hold true, it may still be a non-$v$-th essence sample, although this situation is rare.

\section{Discussion}
\subsection{Polarization}
Apart from enumeration, we have not yet proposed an efficient algorithm to find a particular solution from a general solution. This is the key to whether the EI algorithm can be practically applied. The polarization can be summarized as the following problem.

\begin{problem}
    Given a set of binary classification functions $\{h_{v}^{[n]}(x)  | v \in [1, l_n]\}$, how to flip their local extreme points and expand the values arbitrarily? 
\end{problem}

For example, for a parabolic function $y=a{(x-b)}^2$, by changing the sign of the parameter $a$, we can flip its extreme point, and by increasing the value of parameter $a$, the extreme value will increase. Furthermore, apart from $L(n,-,\Phi)$, $L(u,-,\Phi)$ with $u \in [1,n-1]$ are all atypical homogeneous linear equations. Whether their unique structure is conducive to obtaining a particular solution is also worthy of further discussion.

\subsection{The Output Layer}
For the convenience of calculation and demonstration, we adopted the sigmoid function as the processing unit of the output layer. Although we have demonstrated in Section \ref{subsec: softmax} that when the softmax function is used as the output layer, the corresponding neural network is a weakened model, it is still worth discussing the complete partial differential analysis of the softmax function.

\subsection{Activation Functions}
Our analysis is based on the case where the neural network is a continuous function, that is, the hidden layer neurons use a continuous sigmoid function. If other functions are adopted, especially non-differentiable functions such as the ReLu function, how should the analysis be conducted?

\subsection{Saddle Points}
Our discussion is mainly conducted under the assumption that a sample satisfying the system of equations $\{\frac{\partial}{\partial x_t} h_{v}^{[n]}(x) = 0 | t \in [1, m]\}$ is an extreme point of the binary classification function. For multivariate functions, the fact that all first-order partial derivatives are zero does not necessarily imply that this is an extreme point of the function. It could be a saddle point. It is not a problem if we can find global maxima or minima using the polarization algorithm. Otherwise, this remains a topic worthy of discussion.

\subsection{Alternative Functions}
From our analysis, it can be seen that the strong generalization ability of neural networks depends on the dynamic variability of their function curves, especially the dynamic adjustment of extreme points. Then, can other functions with similar properties provide equally strong generalization ability? For example, the sine function has infinitely many extreme points, and its range is limited to a finite interval. The number of extreme points of a polynomial is positively correlated with its degree. These two seemingly simple functions may have unexpected generalization ability.

\section{Summary}
From a mathematical perspective, we point out the reason why neural networks have strong generalization capabilities, supplementing the shortcomings in the works of \cite{cybenko1989approximation} and \cite{hornik1989multilayer}. We also present the corresponding EI algorithm that is different from the BP algorithm. If there is no effective polarization method, then the EI algorithm can at least become a very important sub-module of the BP algorithm. That is, we can first initialize the parameters using an instance of the EI algorithm's general solution, and then train the neural network using the BP algorithm. If an efficient polarization algorithm is found, then the EI algorithm is expected to become a strong competitor of the BP algorithm, especially when the most ideal model is required.

\acks{I am extremely grateful to my supervisor, Professor Yang Lihua, who provided me with the initial guidance and assistance on the basic research direction when I first started my research journey. Without his guidance, I might have chosen a different research path, thus missing out on this study. Additionally, this research was funded by the Science and Technology Research Project of the Key Areas in Nanhai District, Foshan City (Grant No. 2230032004637). I would like to express my gratitude for this as well.}

\vskip 0.2in
\bibliography{main}

\end{document}